\documentclass[logo, address]{trfr} 
\usepackage{natbib}
\pdfoutput=1
\usepackage{hyperref}
\usepackage{amsmath}
\usepackage{array}   
\usepackage{adjustbox}
\usepackage{wrapfig}
\usepackage{xcolor} 
\usepackage{makecell}
\usepackage{pifont}

\usepackage{latexsym}
\usepackage{enumitem}
\usepackage[utf8]{inputenc}
\usepackage{csquotes}
\usepackage{amsmath,amssymb}
\usepackage{tabularx}
\usepackage{longtable}
\usepackage{array}

\definecolor{ForestGreen}{RGB}{34, 139, 34}

\title{Beyond Pointwise Scores: Decomposed Criteria-Based Evaluation of LLM Responses}

\abstract{
Evaluating long-form answers in high-stakes domains such as law or medicine remains a fundamental challenge. Standard metrics like BLEU and ROUGE fail to capture semantic correctness, and current LLM-based evaluators often reduce nuanced aspects of answer quality into a single undifferentiated score. We introduce \textbf{DeCE}, a decomposed LLM evaluation framework that separates \emph{precision} (factual accuracy and relevance) and \emph{recall} (coverage of required concepts), using instance-specific criteria automatically extracted from gold answer requirements. DeCE is model-agnostic and domain-general, requiring no predefined taxonomies or handcrafted rubrics. We instantiate DeCE to evaluate different LLMs on a real-world legal QA task involving multi-jurisdictional reasoning and citation grounding. DeCE achieves substantially stronger correlation with expert judgments ($r$=0.78), compared to traditional metrics ($r$=0.12), pointwise LLM scoring ($r$=0.35), and modern multidimensional evaluators ($r$=0.48). It also reveals interpretable trade-offs: generalist models favor recall, while specialized models favor precision. Importantly, only 11.95\% of LLM-generated criteria required expert revision, underscoring DeCE’s scalability. DeCE offers an interpretable and actionable LLM evaluation framework in expert domains.
}

\author[1]{Fangyi Yu}
\author[1]{Nabeel Seedat}
\author[2]{Dasha Herrmannova}
\author[2]{Frank Schilder}
\author[1, 3]{Jonathan Richard Schwarz}

\affiliation[1]{Thomson Reuters Foundational Research}
\affiliation[2]{Thomson Reuters Labs}
\affiliation[3]{Imperial College London}

\correspondence{\{first.last\}@thomsonreuters.com}

\begin{document}

\maketitle

\section{Introduction}
As large language models (LLMs) are increasingly deployed in high-stakes, expert settings, such as law, medicine, and finance; their outputs must satisfy demanding requirements: factual accuracy,  citation support, and coverage of domain-specific obligations \citep{lai2024large,wang2024legal,zhang2024evaluation}. However, evaluating such complex, long-form responses remains a fundamental challenge \citep{liang2022holistic, chang2024survey}. Evaluation failures in these domains carry real consequences, including tangible harm, legal liability and erosion of trust in AI systems. To anchor ideas, consider legal question answering, where a lawyer might ask: ``Does a \$2B acquisition of a competitor trigger antitrust filing obligations in California?'' An effective LLM answer must synthesize statutes, regulations, and case law. Subsequent evaluation is hence a fundamentally open-ended, multi-dimensional challenge.

\begin{figure*}[t]
\vspace{-3mm}
    \centering
    \begin{subfigure}[b]{1\textwidth}
        \centering
        \includegraphics[width=\textwidth]{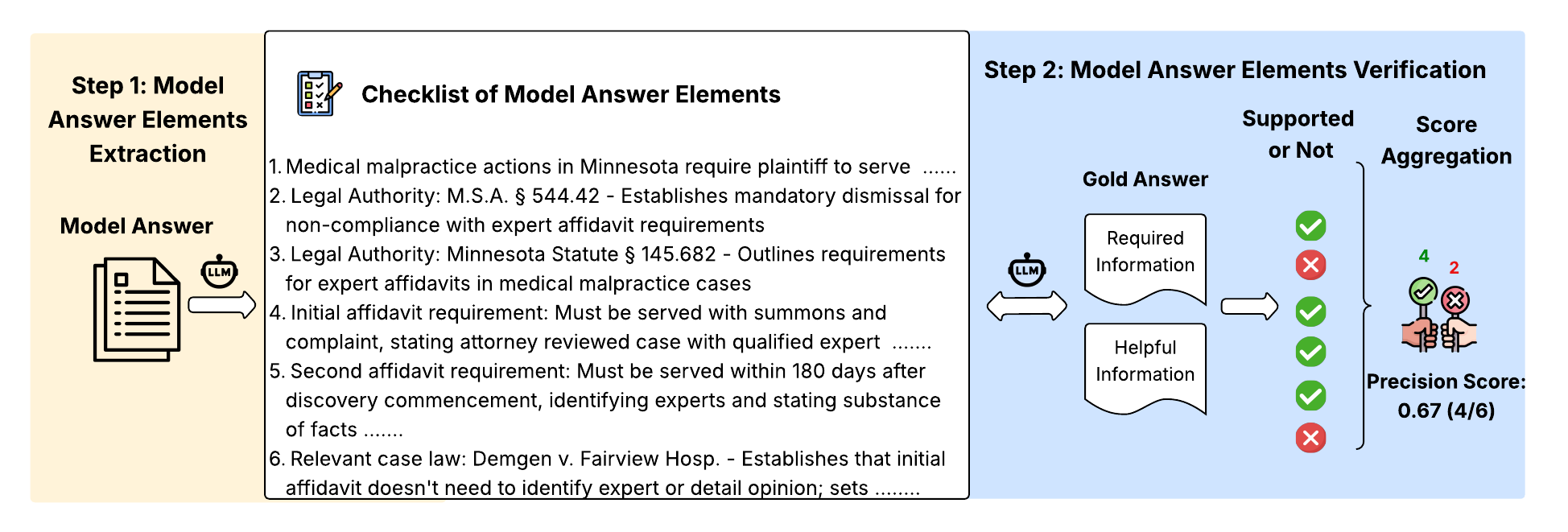}
        \caption{Precision Workflow}
        \label{fig:precision}
    \end{subfigure}
    
    \vspace{0.5cm}  
    
    \begin{subfigure}[b]{1\textwidth}
        \centering
        \includegraphics[width=\textwidth]{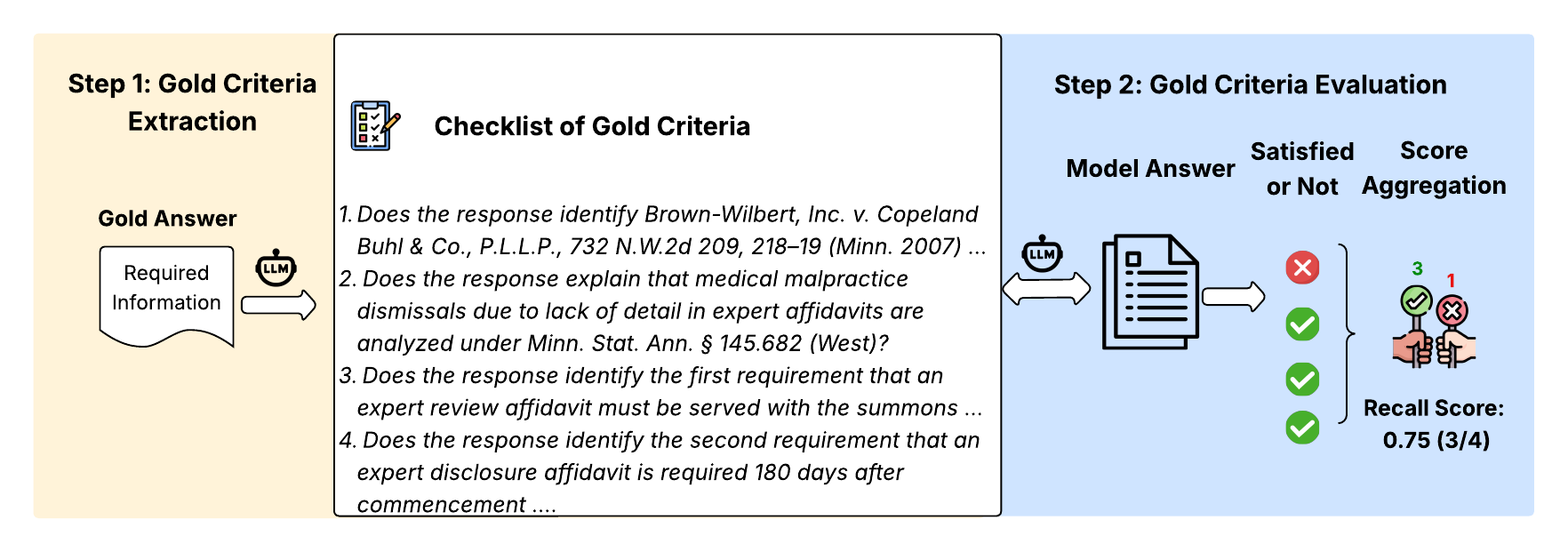}
        \caption{Recall Workflow}
        \label{fig:recall}
    \end{subfigure}
    \caption{Overview of the DeCE evaluation pipeline. 
\textbf{(a)} The precision workflow decomposes the model-generated answer into factual elements, which are then individually verified for factual correctness and relevance against the gold answer. 
\textbf{(b)} The recall workflow extracts evaluation criteria from the gold answer Required Information and checks whether each criterion is satisfied in the model response.
Together, these workflows yield decomposed scores that provide interpretable evaluation signals for expert-domain model evaluation.}
 \vspace{-3mm}
\rule{\linewidth}{.5pt}
 \vspace{-5mm}
    \label{fig:decomposed_autoeval_workflow}
\end{figure*}

Unfortunately, existing evaluation paradigms fail to address this challenge. \textbf{\emph{Human expert reviews}}, though considered the gold standard, are costly and unscalable (e.g., it can take a legal expert around 45 minutes to evaluate one model response by thoroughly checking citations, verifying legal reasoning, and assessing applicability). This approach becomes prohibitively expensive for large-scale evaluation across multiple models and tasks.  \textbf{\emph{Lexical metrics}} like ROUGE \citep{lin-2004-rouge} and BLEU \citep{papineni-etal-2002-bleu} are cheap but correlate poorly with human judgment on complex, knowledge-intensive tasks. \textbf{\emph{LLM-as-a-judge}} has emerged as a promising alternative \citep{zheng2023judging, li2024generation,gu2024survey}, yet many implementations reduce evaluation to a single pointwise score, obscuring actionable insights. Recent multidimensional variants (e.g., GPTScore \citep{fu-etal-2024-gptscore} , G-Eval \citep{liu-etal-2023-g}) add labeled axes (e.g., accuracy, completeness) and show improved alignment in general domains; however, they typically rely on generic, task-agnostic criteria that miss domain-specific obligations and hierarchies central to expert settings like law. \textbf{\emph{Checklist-based approaches}} (e.g., LLM-Rubric \citep{hashemi2024llm},  CheckEval \citep{lee2024checkeval}) increase granularity but require manual rubrics, taxonomies, or seed questions. Hence, while these methods provide greater granularity than naive LLM-as-a-judge, they require substantial manual effort or lack systematic decomposition into dimensions such as precision (factual accuracy, relevance) and recall (coverage), which are essential in expert domains to diagnose specific model behaviors. RAGCHECKER \citep{ru2024ragchecker} advances claim-level precision/recall via bidirectional entailment over extracted claims but treats claims uniformly, overlooking requirement hierarchies. Beyond claim overlap, AQuAECHR \citep{weidinger-etal-2025-aquaechr} and ALCE \citep{gao-etal-2023-enabling} measure precision and recall of citations to assess citation accuracy and coverage in legal contexts, but are limited to citation verification. A comparison of our work vs related works is illustrated in Table \ref{tab:related_work}.

\begin{table*}[!ht]

\vspace{-2mm}
\centering
\caption{Comparison of evaluation frameworks across key dimensions. \checkmark\ indicates full support, {$\triangle$} indicates partial support, and \texttimes\ indicates no support.}
\label{tab:related_work}
\begin{tabular}{l|cccccccccc}
\toprule
\textbf{Method} &
\rotatebox{90}{\textbf{Auto Criteria}} &
\rotatebox{90}{\textbf{Domain-Aware}} &
\rotatebox{90}{\textbf{Inst.-Level Adapt.}} &
\rotatebox{90}{\textbf{Decomp. Eval}} &
\rotatebox{90}{\textbf{Prec/Recall}} &
\rotatebox{90}{\textbf{Manual Taxo. Free}} &
\rotatebox{90}{\textbf{Crit. Interpret.}} &
\rotatebox{90}{\textbf{Mod. Diagnostics}} &
\rotatebox{90}{\textbf{Hierarchy-Aware}} &
\rotatebox{90}{\textbf{Scalable}} \\
\midrule
\textbf{Traditional Metrics} & & & & & & & & & \\
\quad ROUGE / BLEU & \texttimes & \texttimes & \texttimes & \texttimes & \texttimes & \checkmark & $\triangle$ & \texttimes  & \texttimes & \checkmark \\
\quad BERTScore / MoverScore & \texttimes & \texttimes & \texttimes & \texttimes & \texttimes & \checkmark & $\triangle$ & \texttimes  & \texttimes & \checkmark \\
\midrule
\textbf{LLM-as-a-Judge} & & & & & & & & & \\
\quad G-Eval/GPTScore & \texttimes & \texttimes & \texttimes & \checkmark & \texttimes & \checkmark & $\triangle$ & \texttimes  & \texttimes & \checkmark \\
\quad Pointwise Judge (Likert) & \texttimes & \texttimes & \texttimes & \texttimes & \texttimes & \checkmark & \texttimes & \texttimes  & \texttimes & \checkmark \\
\midrule
\textbf{Criteria-Based LLM Judge} & & & & & & & & & \\
\quad CheckEval & $\triangle$ & \checkmark & \texttimes & \checkmark & \texttimes & \texttimes & \checkmark & \texttimes  & \texttimes & $\triangle$ \\
\quad LLM-Rubric & \texttimes & \checkmark & \texttimes & \checkmark & \texttimes & \texttimes & $\triangle$ & \texttimes  & \texttimes & \texttimes \\
\midrule
\textbf{Human Expert Review} & \texttimes & \checkmark & \checkmark & $\triangle$ & \texttimes & $\triangle$ & \checkmark & \checkmark  & \texttimes &  \texttimes \\
\midrule
\textbf{RAGCHECKER} & \checkmark & $\triangle$ & \checkmark & $\triangle$ & \checkmark & \checkmark & \checkmark & \checkmark  & \texttimes & \texttimes \\
\textbf{DeCE (Ours)} & \checkmark & \checkmark & \checkmark & \checkmark & \checkmark & \checkmark & \checkmark & \checkmark & \checkmark & \checkmark \\
\bottomrule
\end{tabular}
\end{table*}

To address these gaps, we propose DeCE, an LLM-based evaluation framework that decomposes evaluation into two interpretable dimensions: precision (factual accuracy and relevance) and recall (coverage of Required Information). DeCE leverages gold-standard answers, typically available in high-stakes domains. Rather than collapsing evaluation into a single opaque score, DeCE automatically extracts instance-specific, domain-aware criteria and uses them to perform a structured precision–recall decomposition that verifies both factual grounding and requirement satisfaction beyond citations.

We include direct comparisons to lexical metrics, pointwise LLM-as-a-judge scoring, GPTScore, G-Eval, and RAGCHECKER in our experiments, and find they align worse with legal expert judgments than DeCE, indicating reduced effectiveness in highly specialized, high-stakes evaluation. See Appendix \ref{app:related_work}  for extended related work.


\textbf{Contributions.}
(1) We propose \textbf{DeCE}, a decomposed criteria-based evaluation framework for LLM evaluation. (2) We show that DeCE aligns significantly better with human expert judgments than standard lexical metrics, holistic LLM-judge scores and multidimensional LLM-as-a-judge baselines, achieving correlation scores up to $r = 0.78$.
(3) We evaluate five diverse frontier LLMs and reveal precision-recall trade-offs.
(4) We demonstrate that DeCE enables fine-grained model behavior analysis, identifying systematic weaknesses of frontier LLMs across jurisdictions and query types.
(5) We validate the reliability of DeCE criteria, finding that only 11.95\% require revision, demonstrating DeCE is scalable and deployable with minimal expert supervision.
\vspace{-0mm}

\section{Decomposed Criteria-Based Evaluation}
\label{sec:dece_formulation}

We introduce \textbf{Decomposed Criteria-Based Evaluation (DeCE)}, a structured LLM-based evaluation framework. Unlike scalar pointwise scoring methods, DeCE decomposes evaluation into two orthogonal, interpretable dimensions: \emph{Precision}: the factual accuracy and relevance of claims made in the model-generated answer and \emph{Recall}: the completeness of the model answer with respect to the Required Information in a gold reference answer.

\subsection{Problem Formulation}

Let each evaluation instance be a tuple \( (q, a_{g}, a_m) \), where \( q \in \mathcal{Q} \) is a question, \( a_{g} \in \mathcal{A}_{g} \) is a gold standard answer with Required Information \( a_{gr} \) and supportive case laws in Helpful Information \( a_{gh} \), and \( a_m \in \mathcal{A} \) is the model-generated answer. 

DeCE computes a decomposed evaluation score:
\[
\text{DeCE}(q, a_{g}, a_m) = (P(q, a_{g}, a_m), R(q, a_{gr}, a_m))
\]
where \( P \) and \( R \) denote precision and recall, derived through automated workflows using an LLM judge comparing \( a_m \) elements against criteria from \( a_{g} \).

\textbf{Motivation and practicality}. Legal analysis prioritizes authorities by hierarchy and direct applicability (e.g., constitutions/statutes > regulations > cases; controlling > persuasive). Expert review therefore distinguishes between requirements that must be satisfied and supporting material that strengthens interpretation. This motivates separating gold answers into “Required” (directly governing) and “Helpful” (supportive/persuasive), which our recall and precision workflows leverage: recall evaluates satisfaction of criteria extracted from \( a_{gr} \), while precision verifies the factual support of elements in \( a_m \) against \( a_{g} \). 

\subsection{DeCE Evaluation Pipeline}

DeCE consists of two self-contained workflows - \textit{Precision Scoring} and \textit{Recall Scoring} - as illustrated in Fig.~\ref{fig:decomposed_autoeval_workflow}. Prompts for each step are detailed in Appendix~\ref{app:DeCE_prompt}, and hyperparameter settings are provided in Appendix~\ref{app:hyperparameters}.

\textbf{1. Precision Scoring}

\noindent \emph{(a) Answer Element Extraction.}  
The model answer \( a_m \) is decomposed into factual elements:
\[
\mathcal{E}_m = \texttt{ExtractElements}(a_m) = \{ e_1, e_2, \dots, e_l \},
\] 
where each \( e_j \) denotes a requirement, principle, or legal authority in \( a_m \).

\emph{(b) Element Verification.}  
Each element $e_j$ is verified against the gold answer $a_{g}$. The precision score is defined as:
$$P(q, a_{g}, a_m) = \frac{1}{|\mathcal{E}_m|} \sum_{j=1}^{|\mathcal{E}_m|} \mathbb{I}\left[ \text{supported}(e_j, a_{g}) \right],$$
where $\mathbb{I}[\text{supported}(e_j, a_{g})] = 1$ if $e_j$ is supported by $a_{g}$, 0 otherwise. This reflects the proportion of model claims grounded in the gold answer.

\textbf{2. Recall Scoring}

\emph{(a) Criteria Extraction.}  
We extract evaluation criteria from Required Information \( a_{gr} \) (excluding additional supportive case laws in \( a_{gh} \) as non-essential for completeness measurement):
\[
\mathcal{C}_g = \texttt{ExtractCriteria}(a_{gr}) = \{ c_1, c_2, \dots, c_k \},
\]
where \( c_i \) represents a query-specific requirement.

\emph{(b) Criteria Satisfaction.}  
The recall score is:
$$R(q, a_{g}, a_m) = \frac{1}{|\mathcal{C}_g|} \sum_{i=1}^{|\mathcal{C}_g|} \mathbb{I}\left[ \text{satisfies}(a_m, c_i) \right],$$
where $\mathbb{I}[\text{satisfies}(a_m, c_i)] = 1$ if $a_m$ satisfies criterion $c_i$, 0 otherwise. This quantifies how fully the model answer covers the essential concepts of the gold answer requirements.

\textbf{\emph{Remark.}} This decomposition enables interpretable evaluation and precise failure attribution, valuable in expert domains.

\section{Experiments}
We evaluate DeCE across multiple dimensions: alignment with expert judgment, model-specific diagnostic insights, and reliability of criteria extraction. We instantiate DeCE on legal question answering, a high-stakes domain where LLMs increasingly assist professionals \citep{LAI2024181}. DeCE is applicable to other domains with structured, expert-authored outputs (clinical QA, financial compliance, scientific summarization, etc.).

\textbf{Dataset.} To assess real-world applicability, we use a professionally curated dataset of 224 English legal QA pairs spanning diverse U.S. state and federal jurisdictions\footnote{Due to proprietary constraints, the data cannot be publicly released.}. While 224 examples may appear modest compared to general-domain QA datasets, each instance reflects high annotation cost in expert domains—often requiring up to an hour of expert time for curation or evaluation.  Hence, this scale is representative of real-world resource constraints in expert domains. Each question is written by expert Attorney Editors and paired with a gold-standard answer that delineates \textit{Required Information} (essential content for recall) and \textit{Helpful Information} (supportive authorities). Relevant legal documents are retrieved using a retrieval system and held constant across all models to mitigate the influence of retrieval. Dataset details are provided in Appendix~\ref{app:dataset}.

\textbf{LLMs evaluated.} We assess five LLMs generating legal answers under standardized retrieval. These include: general-purpose GPT-4o (OpenAI); reasoning models Gemini-2.5-Pro \citep{comanici2025gemini25pushingfrontier} and DeepSeek-R1 \citep{guo2025deepseek}; open-source model Llama-3.1-405B \citep{grattafiori2024llama}; and domain-specific model, Legal Llama-3.1-70B, which we fine-tuned using supervised learning \citep{zhang2024instructiontuninglargelanguage} and Direct Preference Optimization \citep{rafailov2023direct} on legal corpora. All models have over 70 billion parameters\footnote{GPT-4o, DeepSeek-R1, and Llama-3.1-405B were accessed via AWS Bedrock API; Gemini-2.5-Pro via Google Vertex AI API, in accordance with each provider’s terms of service and consistent with their intended use. We fine-tuned Llama-3.1-70B under Meta’s Llama 3 Community License. The finetuning specifics of the Legal Llama-3.1-70B model are out of scope for this paper.}.

\textbf{Evaluation method baselines}. We compare DeCE (using Claude 3.5 Sonnet \citep{claude} as the backbone LLM) against four categories of evaluation methods:
(1) \textbf{Lexical overlap metrics.} ROUGE-L and BLEU: standard overlap metrics that are computationally cheap but correlate weakly with semantic correctness on complex generation tasks.
(2) \textbf{Pointwise LLM-as-a-judge.} Following prior work \citep{zheng2023judging, li2024generation}, we prompt an LLM (Claude 3.5 Sonnet) to assign 0–4 Likert scores relative to the gold answer using a rubric-driven prompt with chain-of-thought \citep{wei2023chainofthoughtpromptingelicitsreasoning}; scores are normalized to a GPA-style 0–4 metric (see prompt in Appendix ~\ref{app:pointwise_prompt}).
(3) \textbf{Multidimensional LLM-as-a-judge.} GPTScore \citep{fu2024gptscore} and G-Eval \citep{liu-etal-2023-g} serve as representative multi-axis judges. For comparability, we: (i) use the same backbone LLM (Claude 3.5 Sonnet), (ii) provide the gold answer as reference , and (iii) report the same dimensions—precision and recall (as in DeCE).
(4) \textbf{Claim-level}. RAGCHECKER \citep{ru2024ragchecker} which computes claim-level precision/recall via bidirectional entailment. We use author-recommended settings and provide the same gold answers and retrieved materials for consistency. We use Claude 3.5 Sonnet for claim extraction for a fair comparison with DeCE.

Extended implementation details and prompts are provided in Appendix ~\ref{app:evaluation}.

\subsection{Does DeCE Align with Human Experts?}\label{sec:exp_alignment}

\noindent\textbf{Goal.}    
Does DeCE's decomposed metrics better align with legal expert judgments compared to the baseline automatic metrics?

\noindent\textbf{Setup.}    
Four U.S. legal experts with 10+ years of practice/academic experience evaluated model responses. For pointwise evaluation, experts assessed all 224 GPT-4o responses using our LLM judge rubric. For decomposed evaluation, experts assessed responses from all five models on 20 randomly selected queries (100 annotations total), evaluating recall (criterion satisfaction) and precision (factual accuracy/relevance). We compute Pearson and Spearman correlations between human judgments and automated metrics. Following established practices in specialized domain evaluation where expert annotation is costly, we employed single-annotator protocol to maximize coverage breadth over agreement assessment.

We use F2 as our primary correlation benchmark, weighting recall over precision for two empirically-motivated reasons: (1) DeCE recall shows stronger correlation with human recall ($r = 0.80$) than DeCE precision with human precision ($r = 0.69$) (see Appendix~\ref{app:correlation_details}), and (2) legal queries often have multiple valid citations beyond gold references, making precision inherently noisy---F2's recall emphasis better captures comprehensive legal reasoning quality while mitigating false negatives from incomplete gold standards.

\begin{table}[htbp]
\centering
\small
\caption{Correlation coefficients between automated metrics and human expert F2. A full correlation comparison is provided in Appendix~\ref{app:correlation_details}.}
\scalebox{0.9}{
\begin{tabular}{l@{\hspace{8pt}}c@{\hspace{8pt}}c@{\hspace{8pt}}c}
\toprule
\textbf{Metric Pair} & \textbf{Pearson} & \textbf{Spearman} & \textbf{P-Value} \\
\midrule
ROUGE-L vs Human & 0.11 & 0.15 & 0.29 \\
BLEU vs Human & 0.12 & 0.13 & 0.13 \\
Point. Judge vs Human & 0.35 & 0.37 & $< 0.05$ \\
GPTScore F2 vs Human & 0.48 & 0.39 & $< 0.05$ \\
G-Eval F2 vs Human & 0.42 & 0.34 &  $< 0.05$  \\
RAGCHECKER vs Human & 0.38 & 0.31 & $< 0.05$  \\
DeCE F2 vs Human & \textbf{0.78} & \textbf{0.76} & $< 0.05$ \\
\bottomrule
\end{tabular}}
\label{tab:correlation}
\end{table}

\noindent\textbf{Analysis.}       
Table~\ref{tab:correlation} shows that ROUGE-L and BLEU exhibit weak correlation with expert assessments (Pearson $r$ = 0.11 and 0.12), confirming their inadequacy for legal responses where semantic correctness and legal reasoning matter more than surface similarity. Pointwise LLM-as-a-judge improves alignment moderately ($r$ = 0.35). Multidimensional LLM-as-a-judge baselines further increase correlation (GPTScore F2: $r$ = 0.48; G-Eval F2: $r$ = 0.42), and RAGCHECKER attains $r$ = 0.38. However, DeCE achieves substantially higher alignment (F2: $r$ = 0.78), indicating that instance-specific, domain-aware criteria and explicit precision–recall decomposition capture expert signals that generic multi-axis rubrics and claim-overlap methods miss.

\textbf{\textcolor{ForestGreen}{Takeaway}.}  
DeCE significantly outperforms existing metrics in aligning with expert judgment, validating its utility as a reliable and low-cost proxy for human evaluation in high-stakes domains.

\subsection{What trade-offs exist across LLMs?}\label{sec:exp_model_analysis}

\noindent\textbf{Goal.}    
We evaluate five different LLMs and aim to highlight insights into their legal answer generation capabilities on the basis of the evaluations.

\noindent\textbf{Setup.}    
We first examine pointwise GPA scores assigned to the generations from the five frontier LLMs. This allows comparison of holistic quality as measured by Likert ratings. We then analyze the decomposed precision and recall distributions produced by DeCE for the same models to investigate trade-offs in factual accuracy and coverage. Note we exclude traditional metrics (ROUGE-L and 
BLEU) from performance analysis due to their 
weak correlation with human judgments.

\begin{figure}[t]
   \vspace{-3mm}
    \centering
    \includegraphics[width=\columnwidth]{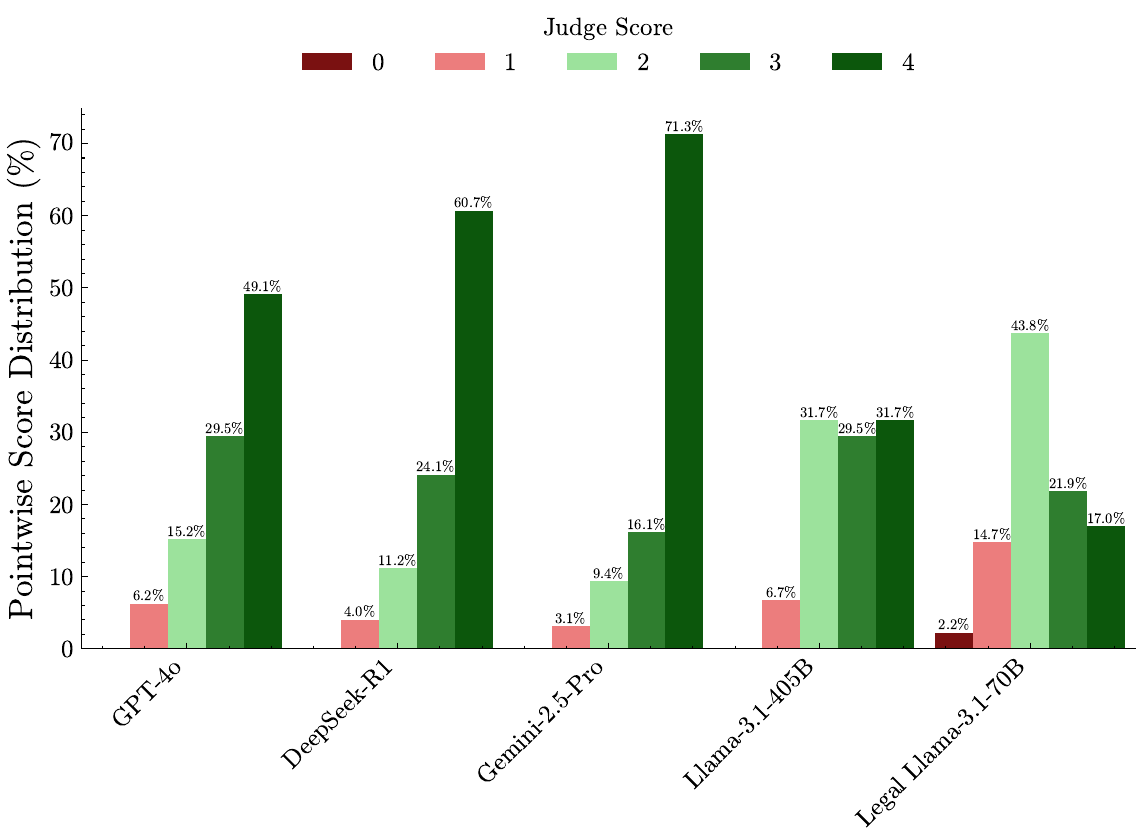}
     \vspace{-7mm}
    \caption{Distribution of pointwise scores (0–4) for each model, judged by Claude 3.5 using rubric-based Likert evaluation. Gemini-2.5-Pro achieves the highest proportion of top-rated responses (71.3\%), while legally fine-tuned Llama-3.1-70B shows lower scores, suggesting model scale may outweigh domain specialization for complex legal reasoning.}
    \label{fig:model_comparison}
     \vspace{-3mm}
\rule{\linewidth}{.5pt}
 \vspace{-7mm}
\end{figure}

\begin{figure}[t]
\vspace{-3mm}
    \centering
    \includegraphics[width=\columnwidth]{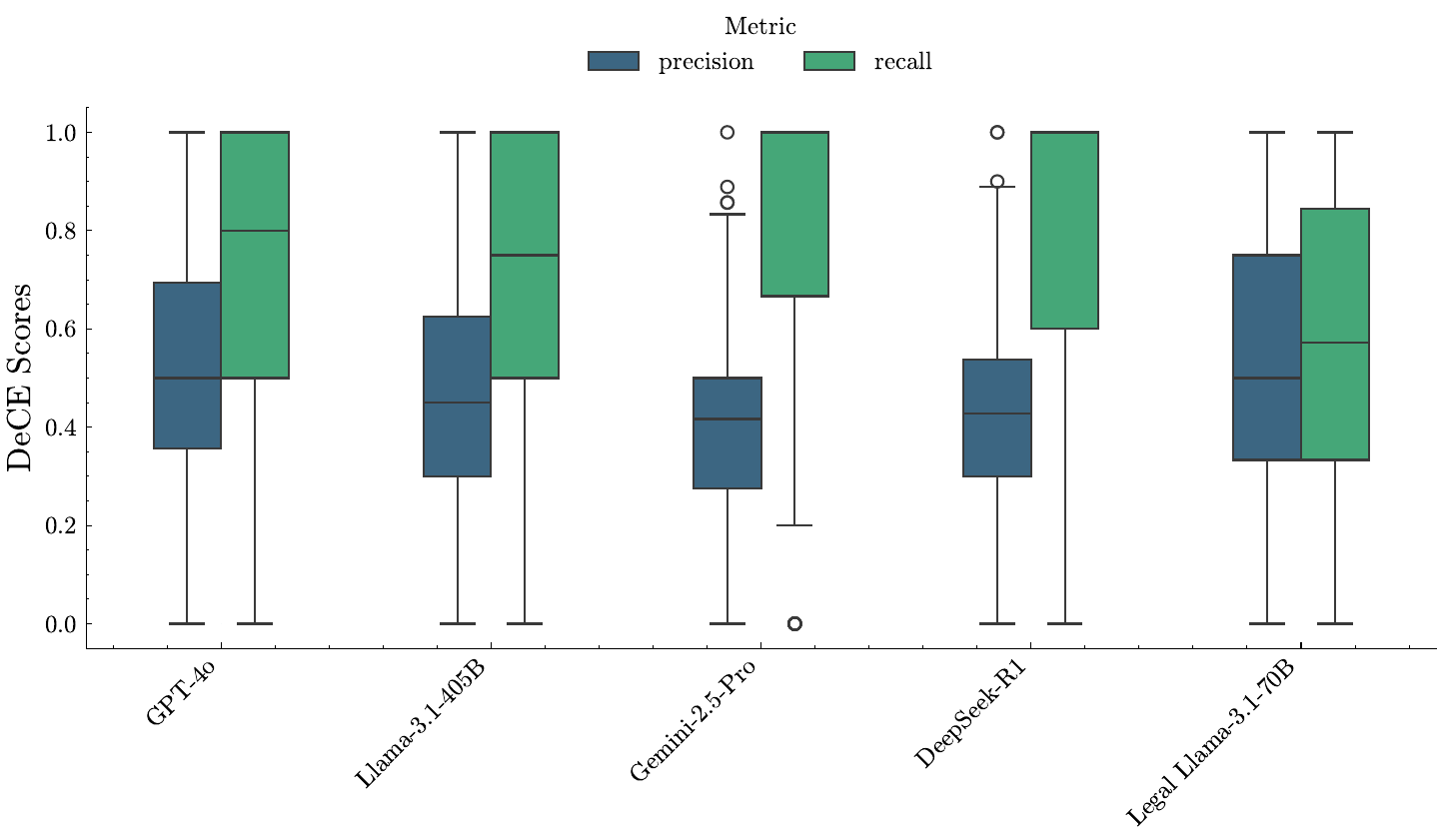}
     \vspace{-7mm}
    \caption{DeCE scores (precision and recall) for each evaluated model. Larger generalist models (e.g., Gemini, GPT-4o) demonstrate stronger recall, while legally fine-tuned models exhibit higher precision, highlighting complementary strengths.}

    \label{fig:pr_distribution}
     \vspace{-3mm}
\rule{\linewidth}{.5pt}
 \vspace{-9mm}
\end{figure}

\noindent\textbf{Analysis.}       
Fig.~\ref{fig:model_comparison} shows Gemini-2.5-Pro achieved the highest GPA (3.56) with 71.3\% ``Excellent'' ratings, followed by DeepSeek-R1 (3.42 GPA, 60.7\% excellent). GPT-4o was balanced (3.21 GPA), while Legal Llama-3.1-70B trailed with only 17.0\% excellent responses, suggesting scale may outweigh domain specialization. 

Fig.~\ref{fig:pr_distribution} reveals insights into precision-recall trade-offs surfaced by our decomposed approach: Gemini-2.5-Pro and DeepSeek-R1 excel at recall (median $\sim$1), indicating strong comprehensiveness, but have lower precision (median $\sim$0.42). In contrast, Legal Llama-3.1-70B achieves the highest precision (median $\sim$0.50) but at the cost of recall. GPT-4o and Llama-3.1-405B perform moderately across both axes. 


\textbf{\textcolor{ForestGreen}{Takeaway}.}  
While GPA scores suggest general performance rankings, DeCE reveals a precision and recall trade-off. Larger models tend to provide more comprehensive (higher recall) but occasionally inaccurate answers (lower precision), whereas smaller or specialized models are more precise but less complete. DeCE permits to diagnose such trade-offs, which holistic metrics obscure.
\subsection{What Insights does DeCE Reveal About Model Behavior?} \label{sec:exp_insights}

\subsubsection{LLM‑Generated Evaluation Criteria Reliability} \label{sec:exp_criteria_reliability}
\noindent\textbf{Goal.} Determine whether an LLM judge can reliably extract evaluation criteria from expert gold answers with minimal human revision.

\noindent\textbf{Analysis.} Expert review of all 979 criteria for the 224-question corpus shows strong reliability.  For 54.5\% of queries the criteria were accepted verbatim.  At the granular level only 11.95\% of individual criteria were modified, 0.7\% were discarded, and 2.0\% were added to capture overlooked nuances.  Revisions fell into three refinement patterns: \textit{specificity calibration} (tightening imprecise wording), \textit{legal authority differentiation} (separating conflated statutory and case‑law requirements), and \textit{case‑law flexibility} (allowing alternative valid precedents).  These low intervention rates indicate that LLM‑generated criteria provide a sound foundation for decomposed evaluation while substantially reducing expert labour. A detailed criteria evaluation analysis is provided in Appendix ~\ref{app:challenge_analysis}.

\textbf{\textcolor{ForestGreen}{Takeaway}.} LLM‑driven criterion extraction is sufficiently accurate for large‑scale use, relegating human experts to a light‑touch validation role.

\begin{figure}[t]
 \vspace{-3mm}
    \centering
    \includegraphics[width=\columnwidth]{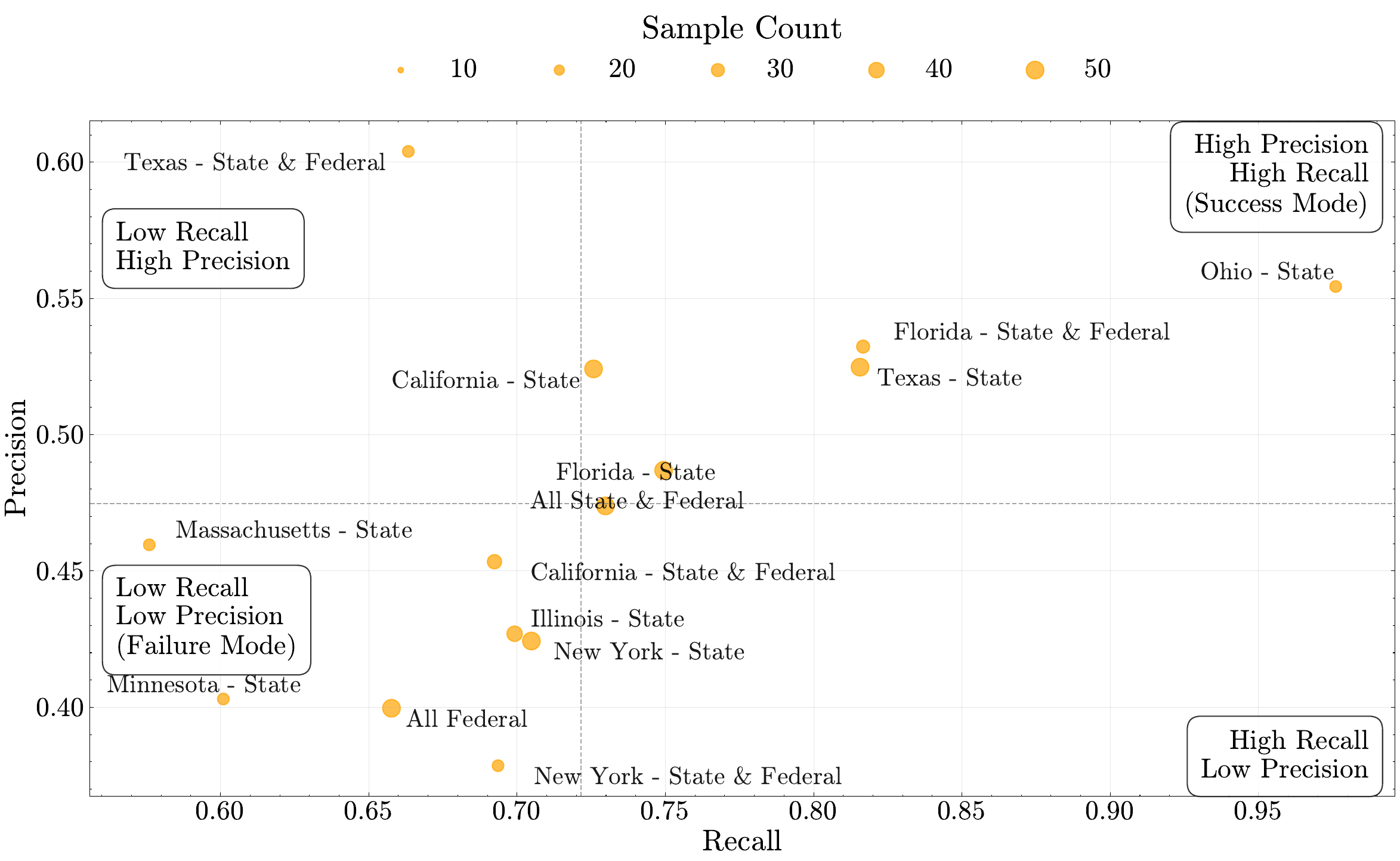}
     \vspace{-7mm}
    \caption{Model performance across jurisdictions insights (precision vs. recall). Ohio State achieves high performance (recall: 0.98, precision: 0.55), while Texas State and Florida State \& Federal show strong balanced performance. New York State \& Federal exhibits low precision (0.38) despite moderate recall, and Minnesota State falls into the failure quadrant with both low precision and recall.}
    \label{fig:jurisdiction_performance}
     \vspace{-3mm}
\rule{\linewidth}{.5pt}
 \vspace{-9mm}
\end{figure}

\subsubsection{Jurisdictional Performance Patterns}
\noindent\textbf{Goal.}   Identify strengths and weaknesses across U.S. jurisdictions using DeCE scores.

\noindent\textbf{Analysis.}      We analyze model performance, observing variation across jurisdiction (Fig.~\ref{fig:jurisdiction_performance}).  Ohio-State queries are handled exceptionally well (average recall 0.98, precision 0.55), while Texas-State and Florida-State show similarly balanced results.  New York-State shows a different failure mode: recall remains moderate but precision drops sharply, signalling unsupported or outdated citations.  Minnesota-State is the most challenging, with deficits on both axes that place all models in the failure quadrant.

\textbf{\textcolor{ForestGreen}{Takeaway}.} Jurisdiction materially affects model behavior.  Outputs concerning New York and Minnesota warrant scrutiny, whereas Ohio‑related queries can be addressed with greater confidence.

\subsubsection{Query‑Type Performance Patterns}
\noindent\textbf{Goal.} Assess which categories of legal questions are well‑handled and which remain problematic.

\begin{figure}[!ht]
    \vspace{-2mm}
    \centering
    \includegraphics[width=\columnwidth]{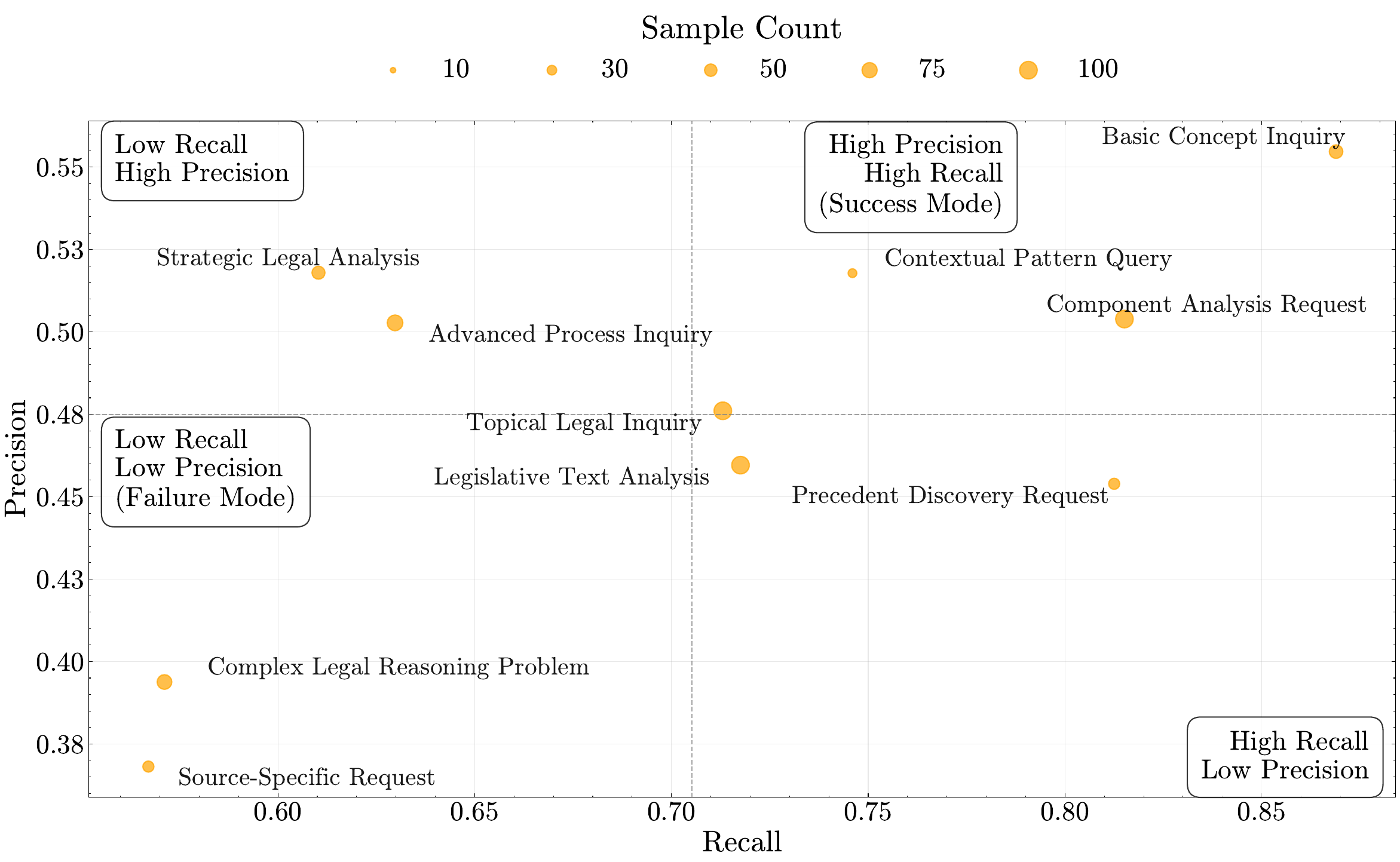}
     \vspace{-7mm}
    \caption{Model performance across query types (precision vs. recall).  Basic concept inquiries achieve optimal performance (recall: 0.87, precision: 0.55), while source-specific requests show the poorest results (recall: 0.57, precision: 0.37). Complex legal reasoning problems consistently challenge all models, highlighting fundamental gaps in legal reasoning capabilities.}
    \label{fig:query_type_performance}
     \vspace{-3mm}
\rule{\linewidth}{.5pt}
 \vspace{-5mm}
\end{figure}

\noindent\textbf{Analysis.} Model performance varies by query type (Fig.~\ref{fig:query_type_performance}).  Basic concept inquries achieve near‑optimal recall (0.87) with respectable precision (0.55).  In contrast, source-specific requests and those requiring legal reasoning capabilities remain difficult: recall ($\pm$0.57) and precision (<0.4). 

\textbf{\textcolor{ForestGreen}{Takeaway}.} Source‑specific and legal reasoning queries expose persistent weaknesses in current systems and should be prioritized for data augmentation and methodological improvement.

\subsubsection{Cross‑Model Challenge Analysis}
\noindent\textbf{Goal.}   Highlight challenges across models.

\noindent\textbf{Analysis.}      Aggregating error slices for GPT‑4o, Gemini 2.5‑Pro, DeepSeek‑R1, and both Llama variants reveals a shared pattern of difficulty.  Massachusetts and Minnesota consistently yield low precision and recall across models, suggesting fundamental challenges rather than architectural limitations.  The same holds for source‑specific requests, and complex legal reasoning problems, which reduce both metrics across models.  

\textbf{\emph{Remark:}} These findings suggest several practical implications: (i) \textit{complementary deployment} strategies could route queries based on jurisdiction and query type to leverage model-specific strengths; (ii) \textit{targeted improvement} efforts could focus on consistently challenging areas like Massachusetts/Minnesota jurisdictions and source-specific requests; and  (iii) \textit{confidence calibration} systems could provide uncertainty indicators for known challenging categories.

\textbf{\textcolor{ForestGreen}{Takeaway}.}
The convergence of failure modes across five diverse models indicates systemic limitations in current LLMs for legal QA. Targeted corpus expansion and human-in-the-loop routing for these challenging cases will likely yield greater benefits than further model scaling alone.

\section{Discussion}

We presented DeCE, a decomposed criteria-based evaluation framework tailored to high-stakes domains like law. By separately assessing \textit{precision} (factual accuracy and relevance) and \textit{recall} (coverage of required concepts) based on automatically extracted criteria from gold answer requirements and elements from model answers, DeCE offers a scalable and interpretable alternative to existing evaluation methods.

Our results demonstrate that DeCE achieves substantially stronger alignment with human expert judgment than traditional lexical metrics ($r = 0.12$) and pointwise LLM-based scoring ($r = 0.35$), with decomposed precision, recall, and F2 reaching correlations of $r = 0.69$, $r = 0.80$, and $r = 0.78$, respectively. Applying DeCE across five frontier LLMs reveals distinct precision–recall trade-offs: larger general models favor recall over accuracy, while legally fine-tuned models yield higher precision but lower completeness. DeCE enables diagnostic insights across jurisdictions (e.g., underperformance in Minnesota and Massachusetts) and query types (e.g., source-specific and legal reasoning requests), revealing challenges common across all LLMs.

Overall, by exposing these nuanced strengths and weaknesses, DeCE enables targeted improvements. The effectiveness of decomposed evaluation establishes a foundation for more sophisticated, domain-sensitive and nuanced evaluation.



\section{Limitations}

\textbf{Dataset Scale and Scope.} Our evaluation dataset, while carefully curated by legal experts, comprises 224 question-answer pairs focused primarily on U.S. jurisdictions. This scale, though sufficient for demonstrating evaluation framework effectiveness and achieving statistically significant correlations with human expert judgments, represents opportunities for broader validation across international legal systems and specialized practice areas.

\textbf{Gold Answer Exhaustiveness Assumption.} Our precision evaluation assumes that gold answers sufficiently cover the space of correct support, particularly with respect to supportive authorities and case law. In practice, legal questions often admit multiple valid lines of reasoning and alternative authorities, and even expert-authored references are not exhaustive. Consequently, precision may underestimate model capability when a response relies on doctrinally valid but unlisted authorities or argument paths. To mitigate this, we outline two complementary extensions that retain DeCE’s decomposed design while better accommodating legitimate multiplicity: 1) Flexible authority matching. During precision verification, treat doctrinally equivalent precedents as supporting when they establish the same controlling rule or holding, even if not explicitly listed in the gold answer (subject to jurisdiction and temporal validity). This reduces false negatives arising from alternative but correct citations. 2) Human-in-the-loop promotion. Introduce a lightweight adjudication pass that reviews frequently recurring, reasonable “false positive” authorities surfaced by DeCE. When validated, these are promoted into the Helpful pool for subsequent runs, expanding coverage without conflating Required content. These extensions preserve interpretability of the precision–recall decomposition—recall continues to target Required criteria, while precision remains a claim-grounding check—yet improve robustness where the gold reference is incomplete.

\textbf{Evaluation Framework Scope.} Our decomposed approach focuses on precision and recall dimensions, which capture completeness and factual accuracy—the primary concerns identified by legal experts in our validation study. While other aspects such as argumentation quality and legal writing style are valuable, our framework addresses the core evaluation challenges that correlate most strongly with expert assessment, providing a solid foundation for future extensions.

\textbf{Model Selection Constraints.} Our evaluation includes representative models across different types (standard, reasoning-optimized, domain-specialized), though broader inclusion of legal-specific larger models with reasoning capability would strengthen generalizability. 

\bibliographystyle{acl_natbib}
\bibliography{main}

\newpage
\appendix

\section*{Appendix}
\section{Extended Related Work}
\label{app:related_work}

\subsection{Human Expert Review}

Human expert evaluation remains the gold standard for specialized domains, providing deep domain knowledge and the ability to adapt evaluation criteria to specific question requirements \citep{lai2024large,wang2024legal}. Expert reviewers naturally perform instance-level adaptation, adjusting their evaluation focus based on the specific legal question, jurisdiction, and context. They can provide interpretable feedback and implicitly assess both precision (accuracy of provided information) and recall (completeness relative to expert expectations). However, human expert review faces fundamental scalability limitations due to cost and expertise requirements \citep{liang2022holistic}.

\subsection{Traditional Evaluation Paradigms}

Early text generation evaluation relied heavily on lexical overlap metrics such as ROUGE \citep{lin-2004-rouge} and BLEU \citep{papineni-etal-2002-bleu}, which measure surface-level similarity between generated and reference texts. These approaches fundamentally fail in expert domains because lexical overlap correlates poorly with human judgments about text quality \citep{chang2024survey}. Human evaluators prioritize semantic meaning, factual accuracy, and domain-specific correctness over word-level similarity.

\subsection{LLM-as-a-Judge}

The emergence of large language models as evaluators represented a paradigm shift \citep{zheng2023judging, li2024generation}. Approaches like G-Eval \citep{liu2023g} and GPTScore \citep{fu2024gptscore} demonstrated that LLMs could achieve better correlation with human judgments than traditional metrics when prompted with evaluation criteria and chain-of-thought reasoning. However, these approaches rely on generic, manually-designed criteria (e.g., relevance, usefulness, clarity) that cannot adapt to domain-specific nuances or individual question characteristics. They treat all evaluation instances identically, missing opportunities for instance-level adaptation and providing minimal interpretability for targeted system improvements.

\subsection{Criteria-Based LLM Evaluation Frameworks}

Recent work has recognized that complex evaluation tasks benefit from decomposition into finer-grained criteria. LLM-RUBRIC \citep{hashemi2024llm} introduced personalized calibration networks that account for individual judge preferences through multi-dimensional frameworks. However, it relies entirely on manually authored rubrics and requires substantial human annotation data to train calibration networks, limiting scalability.

CheckEval \citep{lee2024checkeval} addresses this by using LLMs to generate rubrics, but still requires humans to define evaluation taxonomies and provide seed questions. Critically, CheckEval uses fixed criteria across all instances within a domain, missing instance-level adaptation opportunities, and ultimately collapses scores into single values, preventing systematic decomposition and error analysis.

\subsection{Claim-Based Evaluation: RAGCHECKER}

RAGCHECKER \citep{ru2024ragchecker} proposes a fine-grained evaluation framework for retrieval-augmented generation (RAG) \citep{10.5555/3495724.3496517} that computes precision and recall over extracted atomic claims. It extracts factual claims from both model output and gold answers, then applies bidirectional entailment to determine which claims are supported or missing.

However, RAGCHECKER has two key limitations for expert domains. First, it treats all factual claims as equally important, overlooking task-specific hierarchies critical in high-stakes fields. In legal reasoning, different information types carry distinct importance levels: legal analysis involves navigating complex hierarchies between statutes, regulations, and case law, while requiring careful selection of the most relevant precedents from potentially hundreds of applicable cases. For instance, finding statutes that directly govern a legal issue is more critical than citing tangentially related case law, and among relevant precedents, those most favorable to a client's position hold greater strategic value. RAGCHECKER's flat claim structure cannot capture these domain-specific priorities.

Second, RAGCHECKER relies on rich, free-form human-authored gold answers as reference standards. While appropriate for open-domain QA, such answers are time-consuming to create and difficult to standardize in high-stakes settings, limiting scalability.

\subsection{DeCE: Decomposed Criteria Evaluation}

Our DeCE framework addresses these fundamental limitations through several key innovations:

\textbf{Automatic Domain-Specific Criteria Generation}: Unlike CheckEval's manual dimension definition \citep{lee2024checkeval} or LLM-RUBRIC's manual rubrics \citep{hashemi2024llm}, DeCE automatically extracts evaluation criteria from expert-authored gold answer requirements, which are usually available in expert domains. This approach captures domain expertise while requiring modification in only 11.95\% of cases, enabling scalable deployment across different legal domains without extensive manual taxonomy construction.

\textbf{Instance-Level Adaptive Evaluation}: DeCE generates question-specific evaluation criteria rather than applying fixed criteria across all instances. This addresses a critical gap in existing automated approaches and achieves the instance-level adaptation that only human expert review previously provided, but with systematic scalability.

\textbf{Systematic Precision-Recall Decomposition}: DeCE leverages structured gold answer requirements that specify core informational goals (e.g., ``cite controlling statute'', ``identify most relevant precedent''), inherently capturing domain-specific hierarchies by distinguishing between different information types. This structured approach evaluates whether models address the complex hierarchies inherent in legal reasoning—such as identifying governing statutes versus selecting strategically relevant precedents from hundreds of potential cases—with each requirement weighted according to its role in the overall legal analysis. DeCE thus shifts evaluation from claim overlap toward goal fulfillment and task grounding.

\textbf{Interpretable Performance Analysis}: Unlike holistic LLM judges that provide opaque scores \citep{zheng2023judging}, DeCE offers detailed criterion-level feedback that enables identification of systematic challenges across jurisdictions and query types, providing concrete guidance for legal AI improvement.

\section{Dataset Details}
\label{app:dataset}

\subsection{Jurisdictional Distribution}
Our dataset of 224 legal question-answer pairs spans diverse U.S. jurisdictions with representation proportional to the population of legal practitioners in each region. Consequently, jurisdictions with larger legal communities—such as New York, California, and Texas—appear with greater frequency than less populous states like Wyoming, Montana, or Nebraska. This distribution reflects the natural concentration of legal activity and ensures that evaluated models encounter jurisdictional diversity necessary for broad applicability while maintaining relevance to real-world usage patterns.



\subsection{Data Example}
\label{app:data_sample}

\begin{tcolorbox}[colback=orange!5!white, colframe=orange!75!black, title=Data Example, breakable]

\textbf{Query:}

What constitutes good cause or excusable neglect to be given more time to timely serve complaint in Florida?\newline
\textbf{Search Results:}
\begin{small}
\begin{verbatim}

<document>
[...]
</document>
[other related documents truncated]
\end{verbatim}
\end{small}

\textbf{Gold Answer:}
\begin{small}

\textbf{Required Information:}
Circumstances that may constitute good cause or excusable neglect for failure to timely serve are expansive, and it is 
[other Required Information truncated]

\textbf{Helpful Information:} There is ample case law interpreting what constitutes ``good cause or excusable neglect'' under Rule 1.070(j). Some examples include:

[other related caselaws truncated]

\end{small}

\end{tcolorbox}

\subsection{Taxonomic Categories}
\label{app:taxomony}
The legal queries have been systematically categorized by Attorney Editors according to a comprehensive taxonomy capturing the breadth of legal information needs. Our classification framework encompasses various

distinct categories organized hierarchically based on query complexity and subject matter.

This taxonomic structure enables fine-grained analysis of model performance across different question types and complexity levels.

\section{Evaluation Framework Details}
\label{app:evaluation}

\subsection{Detailed Pointwise Evaluation Rubric}
\label{app:pointwise_rubric}
Our pointwise evaluation employs the following detailed criteria:

\begin{itemize}
    \item \textbf{Irrelevant (0):} Completely incorrect or unrelated to the input and request. The response demonstrates no understanding of the legal question or provides information that is factually wrong or legally inapplicable.
    
    \item \textbf{Poor (1):} Mostly incorrect or largely fails to address the specific request. The response may contain some relevant information but is predominantly inaccurate or misses the core legal issues.
    
    \item \textbf{Fair (2):} Partially correct with noticeable gaps or minor inaccuracies. The response addresses some aspects of the legal question but omits important elements or contains minor factual errors.
    
    \item \textbf{Good (3):} Correct and adequately addresses the request, but lacks some nuanced information. The response covers the main legal points accurately but may miss subtle distinctions or comprehensive coverage.
    
    \item \textbf{Excellent (4):} Fully correct, comprehensive, and directly addresses the specific request. The response demonstrates thorough understanding of the legal issues and provides complete, accurate information.
\end{itemize}

\subsection{Chain of Thought in LLM Judge Prompt}
\label{app:CoT}
\begin{enumerate}
    \item \emph{Query Analysis Phase:} First, carefully analyze the query to understand what legal question is being asked.
    
    \item \emph{Ideal Answer Examination Phase:} Next, examine the ideal answer to understand what a comprehensive response should include. Pay special attention to what information is labeled as \enquote{Required Information} versus \enquote{Helpful Information.} Only penalize for missing \enquote{Required Information.}
    
    \item \emph{Model Answer Review Phase:} Carefully review the model answer to ensure you fully understand its meaning and how it uses citations. Note any ambiguous statements that might affect your evaluation.
    
    \item \emph{Comparative Evaluation Phase:} Next, review the provided reference materials (search results) and compare the model answer against the ideal answer. Evaluate:
    \begin{enumerate}[label=(\alph*)]
        \item Whether all key legal principles from the Required Information in the ideal answer are correctly identified
        \item If citations are properly used to support claims (citations in square brackets [\#] should correspond to relevant paragraph IDs in the search results)
        \item Whether the information provided is accurate and relevant to the question
        \item Any gaps or errors in legal reasoning
        \item If \textsc{all} Required Information from the ideal answer is present (a response can only receive an \enquote{Excellent} grade if it includes all Required Information)
    \end{enumerate}
    
    \item \emph{Issue Identification Phase:} Based on your analysis, identify specific issues using these labels:
    \begin{description}
        \item[Incorrect:] Contains factually or legally inaccurate statements
        \item[Misattribution:] Cited sources do not support the statements they're meant to support
        \item[Missing information:] Lacks essential information in \enquote{Required Information} of the gold answer
        \item[Citation needed:] Contains statements requiring citation but none is provided
        \item[Irrelevant:] Includes information unresponsive to the legal question
        \item[Wrong jurisdiction:] Based on laws from a different jurisdiction than specified
        \item[Repetitive:] Unnecessarily repeats the same legal points multiple times
    \end{description}
    
    \item \emph{Final Grading Phase:} Determine the final grade by weighing the strengths and weaknesses identified.
\end{enumerate}

\subsection{Pointwise Evaluation Prompt}
\label{app:pointwise_prompt}

The prompt template below is domain-agnostic and can be adapted across various specialized fields, including legal, medical, and financial domains.

\begin{tcolorbox}[colback=orange!5!white, colframe=orange!75!black, title=Pointwise Evaluation Prompt Template, breakable]

\begin{small}
You are a [legal/medical/financial] expert evaluating AI-generated responses to specialized queries. Your task is to assess response quality against gold standard answers using established evaluation criteria.\newline

\textbf{Evaluation Framework:}

[The domain-specific rubric. A legal domain rubric is specified in Appendix~\ref{app:pointwise_rubric}]\newline

\textbf{Evaluation Process:}

Follow the structured chain-of-thought methodology below:

[A domain-specific chain-of-thought process; a legal example is provided in Appendix~\ref{app:CoT}]\newline

\textbf{Response Format:}

Provide your complete reasoning process followed by structured output in JSON format:

\begin{verbatim}
{
  "reasoning": "Detailed chain-of-thought analysis...",
  "score": [numerical_grade],
  "justification": "Concise explanation for assigned score"
}
\end{verbatim}

\textbf{Reference Examples:}

Consult the graded response examples below to calibrate your evaluation standards across different quality levels:

[Insert example answers of varying quality for the same query]\newline

\textbf{Evaluation Input:}

Query: \{query\}

Search Results: \{search\_results\}

Gold Standard Answer: \{gold\_answer\}

Model Response: \{model\_response\}

\end{small}
\end{tcolorbox}

\subsection{DeCE Prompt Templates}
\label{app:DeCE_prompt}

DeCE employs four distinct prompt templates corresponding to each evaluation step. All templates are presented below:

\subsubsection{DeCE Criterion Extraction Prompt}
\label{app:dece_criterion_prompt}

The prompt template below is domain-agnostic and can be adapted across various specialized fields to extract evaluation criteria from gold standard answers.

\begin{tcolorbox}[colback=orange!5!white, colframe=orange!75!black, title=Gold Criterion Extraction Prompt Template, breakable]

\begin{small}
You are a [legal/medical/financial] analysis expert tasked with converting comprehensive domain-specific answers into structured assessment criteria.

\textbf{Input:}
\begin{enumerate}
\item A specialized domain query
\item A gold standard answer containing structured information sections (e.g., ``Required Information'' and ``Helpful Information'')
\end{enumerate}

\textbf{Task Specification:}

Create a comprehensive checklist of evaluation criteria by:

\begin{enumerate}
\item \textbf{Extracting mandatory elements} from the Required Information section, including:

[Required elements vary by domain; for the legal domain, the elements are as follows:]
   \begin{itemize}
   \item Key domain principles, requirements, and concepts
   \item Specific conditions, exceptions, or qualifications
   \item Procedural steps or chronological requirements
   \item Evidentiary standards or verification requirements
   \item Domain-specific regulatory or jurisdictional elements
   \item Authoritative sources (cases, regulations, guidelines) and their significance
   \item Analytical frameworks or interpretive approaches
   \end{itemize}

\item \textbf{Incorporating referenced examples} when the primary section explicitly references supplementary information:
   \begin{itemize}
   \item Create criteria that verify inclusion of appropriate illustrative examples
   \item Specify that mentioning any relevant example is sufficient unless otherwise indicated
   \end{itemize}
\end{enumerate}

\textbf{Response Format:}

Return your response in JSON format as follows:

\begin{verbatim}
{
  "gold_criterion": [
    "1: [Description of first required element]",
    "2: [Description of second required element]",
    ...
  ]
}
\end{verbatim}

\textbf{Reference Examples:}

[Insert domain-specific examples demonstrating criterion extraction]\newline

\textbf{Evaluation Input:}

Query: \{query\}

Gold Standard Answer: \{answer\}

\end{small}
\end{tcolorbox}

\subsubsection{Model Answer Evaluation Against Gold Criteria}
\label{app:dece_evaluation_prompt}

The prompt template below evaluates AI-generated responses against gold criteria across various specialized domains.

\begin{tcolorbox}[colback=orange!5!white, colframe=orange!75!black, title=Criterion Evaluation Prompt Template, breakable]

\begin{small}
You are a [legal/medical/financial] expert evaluating an AI-generated response to a specialized domain question. You are provided with ideal answer criteria and an AI-generated response to assess.

\textbf{Input Data:}

\begin{verbatim}
[BEGIN DATA]
***
[Ideal answer criteria]: {gold_criteria}
***
[AI-generated response]: {model_response}
***
[END DATA]
\end{verbatim}

\textbf{Evaluation Task:}

Grade the AI response against each numbered criterion using binary scoring:
\begin{itemize}
\item \textbf{Score 1:} Criterion is satisfied
\item \textbf{Score 0:} Criterion is not satisfied
\end{itemize}

\textbf{Evaluation Guidelines:}

For each criterion, assess satisfaction based on the following rules:

[Evaluation rules vary by domain; the evaluation rules for the legal domain are as follows:]
\begin{enumerate}
\item \textbf{Content matching:} The response addresses the criterion's requirements, regardless of exact wording
\item \textbf{Implicit coverage:} The response implicitly captures the essential elements of the criterion
\item \textbf{Authoritative sources:} When the response cites relevant authorities (cases, regulations, guidelines) with appropriate context, score as satisfied
\item \textbf{Logical equivalence:} Responses stated in negative versus positive form (or vice versa) that encompass criterion elements are acceptable
\item \textbf{Conservative scoring:} When in doubt, assign a score of 0 (not satisfied)
\end{enumerate}

For each criterion, identify and quote specific statements from the response that support your scoring decision and provide clear reasoning.

\textbf{Response Format:}

Provide evaluation in JSON format only. Ensure proper escaping of quotation marks within string fields:

\begin{verbatim}
{
    "scores": [
        [score of 0 or 1 for first criterion],
        [score of 0 or 1 for second criterion],
        ...
    ],
    "reasoning": [
        "[explanation for first criterion scoring]",
        "[explanation for second criterion scoring]",
        ...
    ]
}
\end{verbatim}

\textbf{Example Output:}

\begin{verbatim}
{
    "scores": [0, 1, 0],
    "reasoning": [
        "The response does not mention the required principle.",
        "The response clearly states that ...",
        "The response lacks specific details about ..."
    ]
}
\end{verbatim}

\end{small}
\end{tcolorbox}

\subsubsection{Model Response Element Extraction Prompt}
\label{app:dece_extraction_prompt}

The prompt template below extracts key elements from AI-generated responses across various specialized domains for evaluation purposes.

\begin{tcolorbox}[colback=orange!5!white, colframe=orange!75!black, title=Element Extraction Prompt Template, breakable]

\begin{small}
You are a [legal/medical/financial] analysis expert tasked with extracting all key elements from a model-generated response for evaluation purposes.

\textbf{Input:}
\begin{enumerate}
\item A specialized domain query
\item A model-generated answer to that query
\end{enumerate}

\textbf{Extraction Task:}

Extract and list ONLY the elements that are explicitly present in the model answer. Your job is purely extractive, not evaluative.

\textbf{Critical Guidelines:}
\begin{itemize}
\item Only include information that is EXPLICITLY stated in the model answer
\item Do NOT mention what is missing from the answer
\item Do NOT use phrases like ``no specific sources were cited'' or ``no conditions were provided''
\item Do NOT evaluate the quality, completeness, or accuracy of the answer
\item If an element category has nothing to extract, simply omit it from your response
\end{itemize}

\textbf{Extraction Categories:}

Extract only what IS present in the text, organized by these categories when applicable:

[Categories vary by domain; for the legal domain, the categories are as follows:]
\begin{enumerate}
\item Key domain principles, requirements, and concepts
\item Specific conditions, exceptions, or qualifications
\item Procedural steps or chronological requirements
\item Evidentiary standards or verification requirements
\item Domain-specific regulatory or jurisdictional elements
\item Authoritative sources cited (cases, regulations, guidelines) -- For each source, include:
   \begin{itemize}
   \item The exact name/citation as mentioned in the text
   \item The specific context in which it was cited
   \item The claimed proposition or principle the source supposedly supports
   \item Any direct quotes attributed to the source
   \end{itemize}
\item Interpretive frameworks or analytical approaches
\end{enumerate}

\textbf{Response Format:}

Return your response in JSON format as follows:

\begin{verbatim}
{
  "model_elements": [
    "1: [Description of first element extracted from model answer]",
    "2: [Description of second element extracted from model answer]",
    ...
  ]
}
\end{verbatim}

\textbf{Evaluation Input:}

Query: \{query\}

Model Answer: \{answer\}

\end{small}
\end{tcolorbox}

\subsubsection{Element Verification Against Gold Standard Prompt}
\label{app:dece_verification_prompt}

The prompt template below verifies whether elements from AI-generated responses are supported by gold standard answers across various specialized domains.

\begin{tcolorbox}[colback=orange!5!white, colframe=orange!75!black, title=Element Verification Prompt Template, breakable]

\begin{small}
You are a [legal/medical/financial] expert tasked with verifying if elements from an AI-generated response are supported by a gold standard answer.

\textbf{Input Data:}

\begin{verbatim}
[BEGIN DATA]
***
[Gold standard answer]: {gold_answer}
***
[Elements from an AI-generated response]: {elements}
***
[END DATA]
\end{verbatim}

\textbf{Verification Task:}

Grade the elements from the AI-generated response based on the gold standard answer using binary scoring:
\begin{itemize}
\item \textbf{Score 1:} Element is supported by the gold standard answer
\item \textbf{Score 0:} Element is not supported by the gold standard answer
\end{itemize}

\textbf{Evaluation Guidelines:}

For each numbered element, assess support based on the following rules:

[Evaluation rules vary by domain; for the legal domain, the rules are as follows:]
\begin{enumerate}
\item \textbf{Content alignment:} The gold standard answer addresses the element's content, regardless of exact wording
\item \textbf{Implicit support:} The gold standard answer implicitly captures the essential aspects of the element
\item \textbf{Authoritative validation:} When the gold standard answer cites relevant authorities (cases, regulations, guidelines) that support the element, score as supported
\item \textbf{Logical equivalence:} Gold standard answers stated in negative versus positive form (or vice versa) that encompass the element are acceptable
\item \textbf{Conservative scoring:} When in doubt, assign a score of 0 (not supported)
\end{enumerate}

For each element, identify and quote specific statements from the gold standard answer that support your scoring decision and provide clear reasoning.

\textbf{Response Format:}

Provide evaluation in JSON format only. Ensure proper escaping of quotation marks within string fields:

\begin{verbatim}
{
    "scores": [
        [score of 0 or 1 for first element],
        [score of 0 or 1 for second element],
        ...
    ],
    "reasoning": [
        "[explanation for first element scoring]",
        "[explanation for second element scoring]",
        ...
    ]
}
\end{verbatim}

\textbf{Example Output:}

\begin{verbatim}
{
    "scores": [0, 1, 0],
    "reasoning": [
        "The gold standard answer does not mention this principle.",
        "The gold standard answer clearly states that ...",
        "The gold standard answer lacks support for this element..."
    ]
}
\end{verbatim}

\end{small}
\end{tcolorbox}

\subsection{Hyperparameter Settings}
\label{app:hyperparameters}

All evaluations employ consistent inference parameters to ensure reproducibility and fair comparison across models. We configure Claude 3.5 Sonnet with task-specific hyperparameters optimized for each evaluation component.

\subsubsection{Pointwise Evaluation}
For holistic quality assessment, we configure the LLM judge to generate detailed chain-of-thought reasoning:
\begin{itemize}
    \item \textbf{Temperature:} 0.0 (deterministic output for evaluation consistency)
    \item \textbf{Max tokens:} Maximum input token count + 2,000 (accommodating CoT reasoning, issue identification, and final scoring)
    \item \textbf{Top-p:} 1.0 
\end{itemize}

\subsubsection{DeCE Evaluation Components}

\textbf{Generation Steps (Steps 1 \& 3):} For the extraction of the gold criteria and the answer elements of the model that require various outputs:
\begin{itemize}
    \item \textbf{Temperature:} 0.3 (introducing controlled randomness for comprehensive element identification)
    \item \textbf{Max tokens:} Maximum input token count + 1,000
    \item \textbf{Top-p:} 1.0 
\end{itemize}

\textbf{Verification Steps (Steps 2 \& 4):} For criteria evaluation and model answer element verification requiring deterministic judgments:
\begin{itemize}
    \item \textbf{Temperature:} 0.0 (deterministic output for evaluation consistency)
    \item \textbf{Max tokens:} Maximum input token count + 1,000 (accommodating brief explanations for verification decisions)
    \item \textbf{Top-p:} 1.0 
\end{itemize}

The temperature differentiation reflects the distinct requirements of each evaluation phase: deterministic verification ensures consistent scoring, while moderate randomness in generation steps promotes comprehensive coverage of response elements and gold criteria.

\subsection{Correlation Analysis Details}
\label{app:correlation_details}
\begin{table}[htbp]
\centering
\small
\caption{Correlation coefficients between automated metrics and human expert judgments. DeCE metrics demonstrate significantly higher correlation with human judgments compared with lexical overlap metrics and pointwise LLM-as-a-judge.}
\begin{tabular}{l@{\hspace{8pt}}c@{\hspace{8pt}}c@{\hspace{8pt}}c@{\hspace{8pt}}c}
\toprule
\textbf{Metric Pair} & \textbf{Pearson} & \textbf{Spearman} & \textbf{P-Value} & \textbf{Instance Count} \\
\midrule
ROUGE-L vs Human Point. & 0.34 & 0.33 & $< 0.05$ & 244  \\
BLEU vs Human Point. & 0.20 & 0.17 & $< 0.05$ & 244  \\
ROUGE-L vs Human F2 & 0.11 & 0.15 & 0.29 & 100  \\
BLEU vs Human F2 & 0.12 & 0.13 & 0.13 & 100  \\
Point. Judge vs Human Point. & 0.59 & 0.47 & $< 0.05$ & 244 \\
Point. Judge vs Human F2 & 0.35 & 0.37 & $< 0.05$ & 100 \\
GPTScore Precision vs Human Precision & 0.69 & 0.66  & $< 0.05$ & 100 \\
GPTScore Recall vs Human Recall & 0.56 & 0.48  & $< 0.05$ & 100 \\
GPTScore F2 vs Human F2 & 0.48 & 0.39  & $< 0.05$ & 100 \\
G-Eval Precision vs Human Precision & 0.66 & 0.66  & $< 0.05$ & 100 \\
G-Eval Recall vs Human Recall & 0.53 & 0.46  & $< 0.05$ & 100 \\
G-Eval F2 vs Human F2 & 0.48 & 0.39  & $< 0.05$ & 100 \\
RAGCHECKER Precision vs Human Precision & 0.62 & 0.63  & $< 0.05$ & 100 \\
RAGCHECKER Recall vs Human Recall & 0.39 & 0.34 & $< 0.05$ & 100 \\
RAGCHECKER F2 vs Human F2 & 0.38 & 0.31  & $< 0.05$ & 100 \\
DeCE F2 vs Human Point. & 0.46 & 0.40 & $< 0.05$ & 244 \\
DeCE Precision vs Human Precision & 0.69 & 0.67 & $< 0.05$ & 100 \\
DeCE Recall vs Human Recall & \textbf{0.80} & \textbf{0.82} & $< 0.05$ & 100 \\
DeCE F2 vs Human F2 & \textbf{0.78} & \textbf{0.76} & $< 0.05$ & 100  \\
\bottomrule
\end{tabular}
\label{tab:correlation_all}
\end{table}

\section{Detailed Criteria Evaluation Analysis }
\label{app:challenge_analysis}

\subsection{Complete Criteria Validation Results}
\label{app:criteria_validation_results}

Table~\ref{tab:criteria_validation} provides comprehensive statistics on LLM-generated criteria validation by legal experts.

\begin{table}[htbp]
\centering
\small
\begin{tabular}{lcc}
\toprule
\textbf{Validation Outcome} & \textbf{Count} & \textbf{Percentage} \\
\midrule
No modification required & 855 criteria & 87.33\% \\
Criteria modified & 117 criteria & 11.95\% \\
Criteria rejected & 7 criteria & 0.72\% \\
New criteria added & 20 criteria &  \\

\bottomrule
\end{tabular}
\caption{Expert validation results for LLM-generated evaluation criteria. Among all 979 LLM-generated criteria, only 11.95\% needs modification and an extra 20 criteria need to be added.}
\label{tab:criteria_validation}
\end{table}

\subsection{Detailed Refinement Patterns}
\label{app:detaield_refinement_patterns}
\subsubsection{Specificity Calibration Examples}
\textbf{Original:} ``Does the response identify that Missouri appellate courts apply the substantial evidence standard?''
\textbf{Refined:} `` Does the response identify that Missouri appellate courts apply the substantial evidence standard \textit{when upholding findings of a lower court}?''

\subsubsection{Legal Authority Differentiation Examples}
\textbf{Original:} `` Does the response identify relevant statutory requirements and supporting case law?''
\textbf{Separated into:}
- `` Does the response cite relevant statutory or regulatory authority?''
- `` Does the response include supporting case examples?''

\subsubsection{Case Law Flexibility Examples}
\textbf{Original:} `` Does the response cite Sullivan v. Town of Acton or Canteen Corp. v. City of Pittsfield?''
\textbf{Refined:} `` Does the response cite Sullivan v. Town of Acton, Canteen Corp. v. City of Pittsfield, \textit{or another relevant case establishing municipal liability principles}?''

\section{Responsible NLP Checklist}
\label{app:responsible_checklist}

\setlength\LTleft{0pt}
\setlength\LTright{0pt}

\begin{longtable}{|p{2cm}|p{5cm}|p{7cm}|}
\hline
\textbf{Checklist ID} & \textbf{Question} & \textbf{Answer} \\
\hline
\endfirsthead

\hline
\textbf{Checklist ID} & \textbf{Question} & \textbf{Answer} \\
\hline
\endhead

\hline
\multicolumn{3}{r}{\textit{Continued on next page}} \\
\endfoot

\hline
\endlastfoot

A1 & Did you describe the limitations of your work? & Yes. The limitations are provided in Section 5. \\
\hline
A2 & Did you discuss any potential risks of your work? & Yes. Due to the page limit of the main paper, the risks are provided below.

\textbf{Over-reliance on automated evaluation:} If practitioners become overly dependent on DeCE scores without human oversight, this could lead to deployment of systems that appear to perform well on metrics but fail in real-world scenarios with serious consequences in high-stakes domains like law and medicine. To mitigate this risk, regular validation of evaluation performance by diverse expert panels should be adopted.

\textbf{Potential for gaming:} Once evaluation criteria become known, there's risk that future LLMs could be optimized specifically to perform well on DeCE metrics without genuine improvement in legal reasoning capabilities. To mitigate this risk, we recommend: continuously updating evaluation criteria, using multiple assessment dimensions beyond precision/recall, and implementing adversarial validation with red team testing to prevent models from overfitting to specific patterns. Additionally, maintaining some evaluation criteria as private, implementing multi-stage validation processes, and establishing continuous monitoring systems with expert oversight can help detect suspicious performance patterns and ensure real-world correlation with DeCE scores. \\
\hline
B1 & Did you cite the creators of artifacts you used? & Yes. We cite the creators of artifacts we used throughout the paper, including the models we evaluate, the baseline metrics we compare with, and the judge model: Claude 3.5 Sonnet used as the backbone LLM for DeCE evaluation.

Regarding the dataset, while our legal QA dataset is proprietary and cannot be released, we acknowledge this limitation in the paper
\\
\hline
B2 & Did you discuss the license or terms for use and or distribution of any artifacts? & Yes, terms of use of models are dicussed in LLMs evaluated in Section 3. \\
\hline
B3 & Did you discuss if your use of existing artifact(s) was consistent with their intended use, provided that it was specified? For the artifacts you create, do you specify intended use and whether that is compatible with the original access conditions (in particular, derivatives of data accessed for research purposes should not be used outside of research contexts)? & Yes, the discussion of our use of existing models are discussed in LLMs evaluated in Section 3. \\
\hline
B4 & Did you discuss the steps taken to check whether the data that was collected / used contains any information that names or uniquely identifies individual people or offensive content, and the steps taken to protect / anonymize it? & Yes. We mentioned in the paper the data is proprietary and won't be pulicly released. Due to the page limit of the main paper, the discussions are provided below:

Our legal QA dataset consists of professionally curated attorney-authored questions and expert gold standard answers that are designed for internal evaluation usage. The dataset contains some personally identifiable information (PII) such as judge names, attorney names, plaintiff and defendant names, which are needed for LLM legal use cases. We do not plan to release the data due to these privacy considerations. \\
\hline
B5 & Did you provide documentation of the artifacts, e.g., coverage of domains, languages, and linguistic phenomena, demographic groups represented, etc.? & Yes. While our legal QA dataset is proprietary and cannot be publicly released due to confidentiality constraints, we provide dataset documetation in Appendix B. \\
\hline
B6 & Did you report relevant statistics like the number of examples, details of train / test / dev splits, etc. for the data that you used / created? & Yes. Data description is provided in Dataset in Section 3 and details are provided in Appendix B. \\
\hline
C1 & Did you report the number of parameters in the models used, the total computational budget (e.g., GPU hours), and computing infrastructure used? & Yes, model parameter size is discussed in LLMs evaluated in Section 3. We access all third-party models through APIs, as mentioned in the footnote of the same section. The paper aims to propose a novel LLM-based, domain-agnostic evaluation framework and to share empirical findings on the limitations of state-of-the-art models for legal QA use cases. Proprietary model training strategies are not within the scope of this paper. \\
\hline
C2 & Did you discuss the experimental setup, including hyperparameter search and best-found hyperparameter values? & Yes. Hyperparameter setting for LLM judges are provdided in Appendix C.5.  The paper aims to propose a novel LLM-based, domain-agnostic evaluation framework and to share empirical findings on the limitations of state-of-the-art models for legal QA use cases. Proprietary model training strategies are not within the scope of this paper. \\
\hline
C3 & Did you report descriptive statistics about your results (e.g., error bars around results, summary statistics from sets of experiments), and is it transparent whether you are reporting the max, mean, etc. or just a single run? & Yes. We report descriptive statistics for our results with full transparency about our experimental setup:

\textbf{Statistical Reporting:}

Correlation coefficients: We report both Pearson and Spearman correlation coefficients between automated metrics and human expert judgments (Table \ref{tab:correlation}).

Performance distributions: We present detailed score distributions across models using box plots (Figure \ref{fig:pr_distribution}) and percentage breakdowns (Figure \ref {fig:model_comparison}).

Aggregate statistics: We report mean scores and performance ranges across different jurisdictions (Figure \ref{fig:jurisdiction_performance}) and query types (Figure \ref{fig:query_type_performance}).

\textbf{Experimental Design:} All reported results are based on single run experiments considering we use temperature=0.0 for all verification steps in DeCE to ensure consistent scoring. \\
\hline
C4 & If you used existing packages (e.g., for preprocessing, for normalization, or for evaluation, such as NLTK, Spacy, ROUGE, etc.), did you report the implementation, model, and parameter settings used? & Yes. Due to the page limit of the main paper, the package versions are provided below:

nltk: 3.9.1

rouge: 1.0.1 

transformers: 4.52.4 \\
\hline
D1 & Did you report the full text of instructions given to participants, including e.g., screenshots, disclaimers of any risks to participants or annotators, etc.? & Partially. We provide a detailed methodology and evaluation criteria, but cannot release the complete annotation instructions due to proprietary constraints.

\textbf{What we do provide:}

Detailed evaluation rubrics: Complete pointwise evaluation criteria (0–4 scale) with specific definitions for each score level (Appendix C.1).

Chain-of-thought methodology: A full six-phase evaluation process, including Query Analysis, Ideal Answer Examination, Model Answer Review, Comparative Evaluation, Issue Identification, and Final Grading (Appendix C.2).

Comprehensive prompt templates: All four DeCE prompt templates for criterion extraction, model evaluation, element extraction, and verification (Appendix C.4).

\textbf{What we cannot provide:}

Complete annotation instructions, which contain proprietary legal evaluation frameworks developed for internal use.

Our provided methodology, rubrics, and prompt templates contain sufficient detail to reproduce the evaluation approach and serve as a guideline for other high-stakes domains, as one purpose of the paper is to propose our novel, domain-agnostic evaluation framework.

Domain experts can adapt our framework using the comprehensive templates and evaluation criteria we provide.

\textbf{Justification:}
This approach balances transparency with proprietary constraints while providing sufficient methodological detail for reproducibility in similar expert domains. \\
\hline
D2 & Did you report information about how you recruited (e.g., crowdsourcing platform, students) and paid participants, and discuss if such payment is adequate given the participants’ demographic (e.g., country of residence)? & Yes. Due to the page limit of the main paper, the discusssion is provided below:

The legal experts who participated in our evaluation are employees of our institution rather than external recruited participants. As institutional employees, their participation in this research evaluation was part of their professional duties rather than separate recruitment with additional compensation.

All experts are based in the United States, appropriate for our U.S. legal QA dataset spanning diverse jurisdictions. \\
\hline
D3 & Did you discuss whether and how consent was obtained from people whose data you’re using/curating? & Yes. Due to the page limit of the main paper, the discussion is provide below:

The human evaluation annotations were conducted by legal experts who are employees of our institution. As institutional employees, their participation in annotation tasks was part of their professional duties rather than external data collection requiring separate consent procedures. \\
\hline
D4 & Was the data collection protocol approved (or determined exempt) by an ethics review board? & Yes, our legal QA dataset use was reviewed and approved by our institution's internal legal department.

The legal professionals who participated in annotation and evaluation tasks are employees of our institution, operating under established professional responsibility frameworks.

All data handling and research activities were conducted in accordance with our institution's internal policies for proprietary data use. \\
\hline
D5 & Did you report the basic demographic and geographic characteristics of the annotator population that is the source of the data? & Yes, as discussed in Section 3.1. \\
\hline
E1 & If you used any AI assistants, did you include information about your use? & Yes. Due to the page limit of the main paper,
the discussion is provide below:

We acknowledge the use of large language models for manuscript refinement and code development assistance during the preparation of this work. All conceptual contributions, experimental design, analysis, and conclusions remain the authors' original work. \\
\hline

\end{longtable}
\clearpage
\end{document}